\documentclass[letterpaper]{article} 
\usepackage{aaai2027}  

\usepackage[hyphens]{url}  
\usepackage{graphicx} 
\usepackage{natbib}  
\usepackage{caption} 
\usepackage{algorithm}
\usepackage{algorithmic}

\usepackage{newfloat}
\usepackage{amssymb}
\usepackage{listings}
\DeclareCaptionStyle{ruled}{labelfont=normalfont,labelsep=colon,strut=off} 
\floatstyle{ruled}
\newfloat{listing}{tb}{lst}{}
\floatname{listing}{Listing}

\usepackage{booktabs}\usepackage{xcolor}

\usepackage{booktabs}

\title{DialogMem: Evaluating Dialogue-Grounded Embodied Memory \\ in 3D Environments}
\author{
    SeongJun Jeong\textsuperscript{\rm 1},
    Junghyun Kim\textsuperscript{\rm 1},
    Gi-Cheon Kang\textsuperscript{\rm 2}\corresponding,
    Byoung-Tak Zhang\textsuperscript{\rm 1}\corresponding
}
\affiliations{
    \textsuperscript{\rm 1}Seoul National University\\
    \textsuperscript{\rm 2}Ajou University\\
    \{sjjeong, jhkim\}@bi.snu.ac.kr, gckang@ajou.ac.kr, btzhang@bi.snu.ac.kr
}

\nocopyright 

\begin{document}

\maketitle

\begin{abstract}
Embodied agents in human-centered environments require memory that integrates what they observe with what humans tell them over time.
However, existing embodied memories are built primarily from visual experience, leaving open how dialogue with humans can be grounded in the objects encountered throughout that experience.
We introduce \textsc{DialogMem}, a two-phase benchmark in which agents build memory from time-aligned streams of posed RGB-D observations and dialogue, and later use it for instance-level object navigation and question answering (QA).
\textsc{DialogMem} provides 142 episodes across 27 real-world 3D scenes, with 14{,}193 QA pairs and 11{,}890 navigation instructions.
We evaluate nine memory representations within a unified framework of memory formation, evidence retrieval, and reasoning for task outputs.
In experiments, appending dialogue to visual memory at evaluation time substantially improves QA but yields negligible navigation gains, suggesting that relevant dialogue is present in memory but not grounded to specific objects.
Controlled diagnostics show that navigation improves with link quality but falls short of ground-truth object-level memory even under perfect linking. Retrieving the correct object entry and recovering its spatial location remain further bottlenecks.
These results identify three bottlenecks for dialogue-grounded embodied memory: grounding and organizing dialogue at object-level entries during formation, retrieving the correct entry, and localizing it accurately.
\end{abstract}


\section{Introduction}

Humans build memories of their environments through extended experience.
They remember which objects they observed, when and where they encountered them, and what others told them about those objects.
Embodied agents operating in human-centered environments need a similar capacity: integrating perceptual evidence, trajectory context, and dialogue-derived information into persistent memory.

Embodied memory research has made rapid progress in turning long streams of visual experience into persistent representations. 
These representations support navigation~\cite{wani2020multion, khanna2024goat}, question answering~\cite{majumdar2024openeqa, hu20253dllm}, and object search~\cite{yadav2025findingdory}, and are often queried at evaluation time by a vision-language model (VLM)~\cite{bai2025qwen3, hurst2024gpt}.
Yet most existing benchmarks~\cite{majumdar2024openeqa, khanna2024goat, yadav2025findingdory, hu20253dllm} and methods~\cite{gu2024conceptgraphs, yang20253d} build memory primarily from RGB-D observations and agent poses, without explicitly retaining dialogue-derived information. We therefore ask: \textit{when experience also includes dialogue, how should embodied memory be organized so that a VLM can both recall what was said and localize the objects it refers to?}

To address this, we introduce  \textsc{DialogMem}, a benchmark for evaluating how well agents build memory from time-aligned streams of posed RGB-D observations and dialogue, and later use it for instance-level object navigation and question answering (QA).
Navigation and QA probe two distinct aspects of dialogue-grounded memory: spatial grounding of referenced objects and semantic access to what was said about them.
\textsc{DialogMem} provides 142 episodes across 27 real-world 3D scenes, with 14,193 QA pairs and 11,890 navigation instructions.
Each episode is split into two phases.
In Phase~1, the agent traverses a 3D environment while receiving time-aligned streams of posed RGB-D observations and dialogue turns, forming memory over an extended multimodal experience.
In Phase~2, the resulting memory is queried by navigation instructions and questions that draw on three types of memory content: \emph{perceptual} content capturing object appearance and spatial relations, \emph{episodic} content from the trajectory, and \emph{social} content from dialogue. Each query refers to a target object using one or more of these content types. Both tasks require identifying the referenced object. Navigation further requires recovering its spatial location, while QA requires retrieving information about the target.

We evaluate nine memory representations on \textsc{DialogMem}, spanning raw stream storage, structured visual memories~\cite{gu2024conceptgraphs, yang20253d}, and dialogue-augmented variants that either append dialogue at evaluation time or link dialogue turns to memory entries during formation. 
All representations use the same VLM reasoner~\cite{bai2025qwen3}, enabling controlled comparison of memory representation and retrieval. 
Across all memory representations, navigation and QA exhibit a clear asymmetry: dialogue substantially improves QA but yields negligible navigation gains.
Neither evaluation-time appending nor simple formation-time linking reliably grounds dialogue-derived information in the objects it refers to. 
A controlled sweep over link quality shows that navigation improves as links become more accurate, yet even perfect links fall short of ground-truth object-level memory.
Controlled diagnostics further reveal substantial bottlenecks in retrieving the correct object entry and recovering its spatial location. Changing the reasoner does not close the gap.
These findings motivate object-centered memories that ground and organize what agents observe and what they are told around the correct object entries during memory formation.
\textsc{DialogMem} provides a testbed for measuring progress on this goal, together with the retrieval and localization bottlenecks it exposes.

\paragraph{Contributions.}
\begin{itemize}
    \item We formulate dialogue-grounded embodied memory as forming persistent memory from posed RGB-D and dialogue streams, retrieving relevant evidence, and reasoning over it for embodied tasks.
    \item We introduce \textsc{DialogMem}, the first benchmark that treats dialogue as a persistent memory source jointly grounded with visual experience and evaluates the resulting memory through both navigation and QA.
    \item We evaluate nine memory representations under a common reasoning framework and introduce controlled link-quality and oracle probes that decompose failure into dialogue--object grounding, entry retrieval, and 3D localization.
    
\end{itemize}
\section{Related Work}

\noindent\textbf{Dialogue in Embodied Tasks.}
Language has been widely used as an interface for embodied agents~\cite{thomason2020vision, gao2022dialfred, padmakumar2022teach}, including navigation instructions, task guidance, clarification, and human-agent collaboration.
However, these settings typically treat dialogue as task-time context rather than as content to be retained across tasks.
In contrast, \textsc{DialogMem} treats dialogue as a memory source during memory formation, requiring agents to ground dialogue-derived object information in perceptual and spatial experience for later navigation and question answering.

\noindent\textbf{Embodied Memory Benchmarks.}
Prior embodied memory benchmarks have made important progress in testing memory over long-horizon visual experience. Evaluation targets include object locations~\cite{wani2020multion, khanna2024goat}, spatial layouts~\cite{pasukonis2022evaluating, pleines2025memory, zhu2023excalibur}, and temporal observation history~\cite{majumdar2024openeqa, min2024situated, hu20253dllm, yadav2025findingdory}.
The input streams in these benchmarks are largely limited to visual observations and poses, leaving out dialogue-derived object information such as aliases, usage patterns, and corrections.
Consequently, these benchmarks primarily evaluate perceptual and episodic memory, leaving social memory largely unexplored. 
\textsc{DialogMem} treats dialogue as a persistent memory source alongside visual experience, thereby extending embodied memory evaluation to include social memory.
Table~\ref{tab:benchmark_comparison} summarizes this distinction.

\noindent\textbf{3D Scene Representations as Embodied Memory.}
Recent 3D scene representations, such as semantic maps~\cite{huang2023visual}, 3D scene graphs~\cite{gu2024conceptgraphs}, and dense 3D representations~\cite{yang20253d, hong20233d}, can serve as embodied memory for downstream tasks by organizing past RGB-D observations into persistent spatial structures.
Recent systems often pair such representations with VLM reasoners~\cite{bai2025qwen3, hurst2024gpt}, using retrieved visual or spatial evidence for tasks such as object localization and navigation.
However, these representations are typically constructed from visual observations alone, leaving open how dialogue-derived object information should be incorporated into object-level memory.

\begin{figure*}[ht!]
    \centering
    \includegraphics[width=\linewidth]{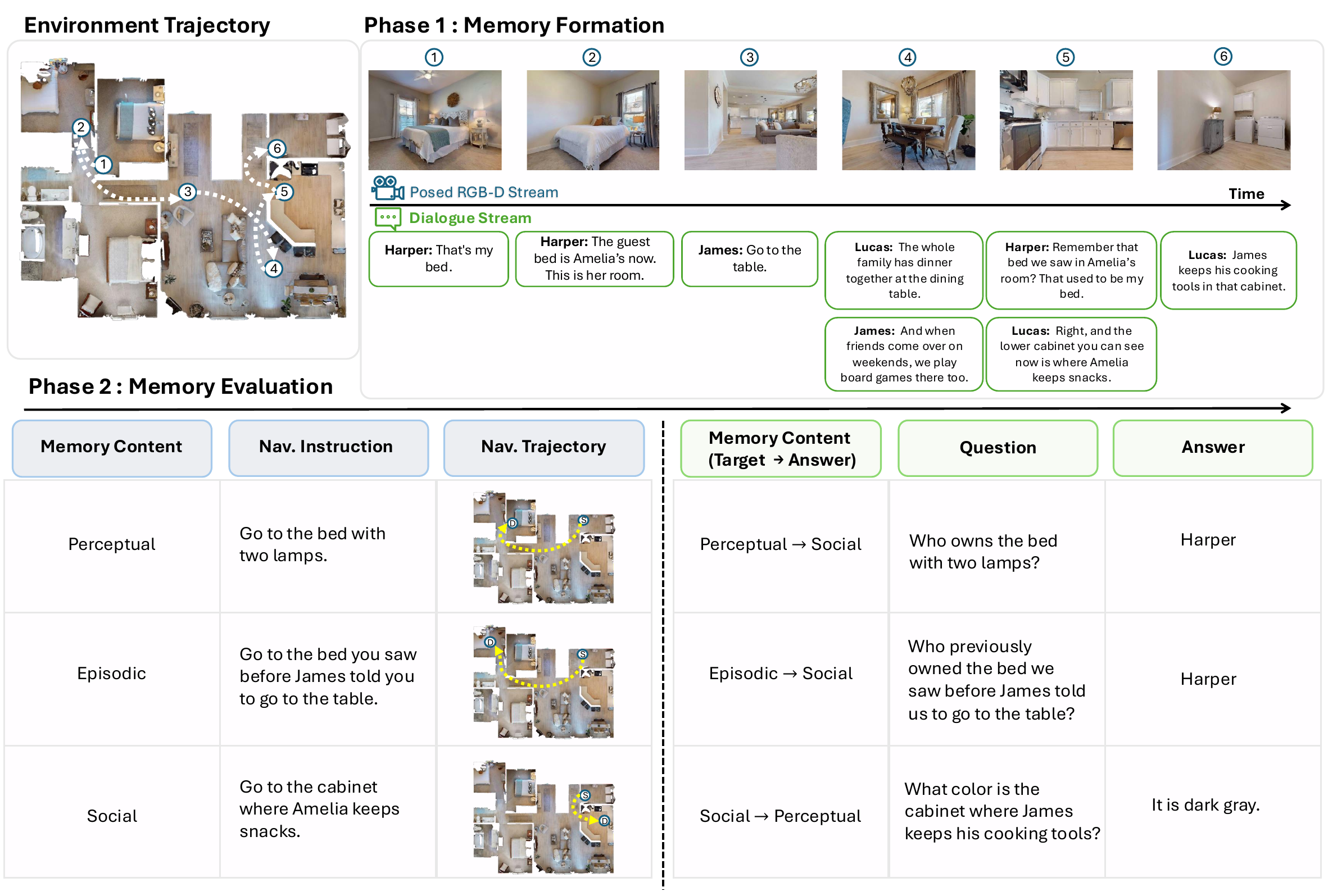}
    \caption{\textbf{Task overview with representative examples.} 
        In Phase 1, the agent receives time-aligned streams of posed RGB-D and dialogue. In Phase 2, the resulting memory supports navigation and QA tasks drawing on three memory contents: \textit{perceptual} (object appearance and spatial relations), \textit{episodic} (trajectory experience), and \textit{social} (dialogue-derived facts). For QA, the arrow notation shows which memory content identifies the target (before the arrow) and which provides the answer (after the arrow). All Phase 2 examples share the same Phase 1 stream.}
    \label{fig:overview}
\end{figure*}

\begin{table*}[t]
\centering
\small
\setlength{\tabcolsep}{2pt}
\begin{tabular}{lcccccccc}
\toprule
\textbf{Benchmark} 
& \textbf{Real.} 
& \textbf{Memory} 
& \textbf{Isolated} 
& \textbf{Dialogue} 
& \textbf{Dialogue} 
& \textbf{Perceptual} 
& \textbf{Episodic} 
& \textbf{Social} \\
& \textbf{3D} 
& \textbf{Eval.} 
& \textbf{Phase} 
& \textbf{Present} 
& \textbf{as Memory} 
& 
& 
& \\
\midrule
CVDN~\cite{thomason2020vision}        & \checkmark & -- & -- & \checkmark & -- & -- & -- & -- \\
DialFRED~\cite{gao2022dialfred}    & --         & -- & -- & \checkmark & -- & -- & -- & -- \\
TEACh~\cite{padmakumar2022teach}       & --         & -- & -- & \checkmark & -- & -- & -- & -- \\
\midrule
MemoryMaze~\cite{pasukonis2022evaluating}  & --         & \checkmark & -- & -- & -- & \checkmark & -- & -- \\
MemoryGym~\cite{pleines2025memory}   & --         & \checkmark & -- & -- & -- & \checkmark & -- & -- \\
Excalibur~\cite{zhu2023excalibur}   & --         & \checkmark & -- & -- & -- & \checkmark & -- & -- \\
MultiON~\cite{wani2020multion}     & \checkmark & \checkmark & -- & -- & -- & \checkmark & -- & -- \\
OpenEQA~\cite{majumdar2024openeqa}     & \checkmark & \checkmark & -- & -- & -- & \checkmark & -- & -- \\
GOAT-Bench~\cite{khanna2024goat}  & \checkmark & \checkmark & -- & -- & -- & \checkmark & -- & -- \\
SIF~\cite{min2024situated}         & -- & \checkmark & -- & -- & -- & \checkmark & \checkmark & -- \\
3DMEM-Bench~\cite{hu20253dllm} & \checkmark & \checkmark & -- & -- & -- & \checkmark & \checkmark & -- \\
FindingDory~\cite{yadav2025findingdory} & -- & \checkmark & \checkmark & -- & -- & \checkmark & \checkmark & -- \\
\midrule
\textsc{DialogMem} (Ours)
             & \checkmark & \checkmark & \checkmark 
             & \checkmark & \checkmark 
             & \checkmark & \checkmark & \checkmark \\
\bottomrule
\end{tabular}
\caption{
\textbf{Comparison with embodied dialogue and embodied memory benchmarks.}
``Real. 3D'' denotes whether the benchmark uses real-world scanned 3D environments.
``Isolated Phase'' denotes whether memory formation is separated from later evaluation.
``Dialogue as Memory'' denotes whether dialogue is used as a persistent memory source.
Perceptual, Episodic, and Social indicate the memory content evaluated by downstream tasks.
}
\label{tab:benchmark_comparison}
\end{table*}

\section{Embodied Memory Formulation}
\label{sec:formulation}

We formalize embodied memory to provide a unified framework for describing and comparing the memory representations evaluated in this work.

\paragraph{Input streams and memory.}
We consider an embodied agent operating in a 3D environment over timesteps $t = 1, \ldots, T$.
At each step, the agent receives an input tuple:
\begin{equation}
s_t = (I_t, p_t, D_t),
\end{equation}
where $I_t$ is an RGB-D frame, $p_t \in SE(3)$ is the agent pose, and $D_t$ is the dialogue content aligned with time $t$, when available.
The agent constructs or updates an internal state $\mathcal{M}_t$, referred to as its \emph{embodied memory}, updated incrementally by an update function $\mathcal{U}$:
\begin{equation}
\mathcal{M}_t = \mathcal{U}(\mathcal{M}_{t-1}, s_t).
\end{equation}

\paragraph{Retrieval and reasoning.}
At evaluation time, given a query $q$, a retrieval function $\mathcal{R}$ selects task-relevant evidence $z = \mathcal{R}(\mathcal{M}_T, q)$.
A reasoner $\mathcal{F}$ then produces the output $y = \mathcal{F}(z, q)$. 
The form of $y$ depends on the task, such as a target location for navigation or an answer for question answering.
This formulation is agnostic to how $\mathcal{M}$ is represented, allowing it to take the form of either raw observation streams or structured visual representations, such as scene graphs and 3D spatial representations.


\section{DialogMem}
\label{sec:benchmark}

\textsc{DialogMem} instantiates the preceding formulation in real-world scanned 3D environments, where agents must integrate perceptual, episodic, and social memory content for downstream navigation and question answering.

\subsection{Task}

\paragraph{Overview.}
The two-phase structure separates memory formation from memory use, so that downstream performance reflects what the agent has retained rather than what it can infer from the current observation (see Figure~\ref{fig:overview}).

\paragraph{Phase 1: Memory formation.}
The agent receives posed RGB-D observations and dialogue streams over an extended traversal.
The goal is to construct a memory $\mathcal{M}_T$ that covers three types of memory content.
\textbf{Perceptual} content captures object appearance and spatial
relations, originating from RGB-D observations.
\textbf{Episodic} content comes from the trajectory and captures when and how often each object was observed.
An instruction such as \textit{``go to the sofa''} further defines a navigation segment, associating objects with segments and supporting references such as \textit{``the chair you saw on the way to the sofa.''}
\textbf{Social} content comes from dialogue and captures human-provided object information, including ownership, aliases, usage patterns, and updates or corrections.

\paragraph{Phase 2: Memory evaluation.}
Phase~2 evaluates whether the agent can retrieve and use the memory content formed in Phase~1.
It consists of two task types grounded in the same Phase~1 episode: navigation and question answering.
In \textbf{navigation}, starting from its Phase~1 endpoint, the agent receives a natural language instruction that identifies a target object and navigates to its location, e.g., \textit{``Go to the cabinet where Amelia keeps snacks.''}
In \textbf{question answering}, the agent first identifies a target object using Phase~1 memory evidence and then answers a question about that object, e.g., \textit{``Who owns the bed with two lamps?''}

\subsection{Benchmark Construction}

\paragraph{Environment and metadata.}
We build \textsc{DialogMem} on 27 scenes from the HM3D validation split~\cite{ramakrishnan2021habitat}, using object annotations from GOAT-Bench~\cite{khanna2024goat}.
\textsc{DialogMem} is an evaluation-only benchmark: following prior embodied QA evaluation practice~\cite{majumdar2024openeqa}, all baselines use pretrained models without training or fine-tuning on \textsc{DialogMem}.
For each scene, we use an LLM (GPT-4o~\cite{hurst2024gpt}) to generate scene-specific resident profiles. Each profile specifies a resident's name, relationship to others, and speaking style, giving dialogue generation a consistent set of speakers.
For each annotated object, we construct perceptual memory content by parsing GOAT-Bench descriptions into attributes and spatial relations. We construct social memory content with an LLM conditioned on the resident profiles.
Social memory content includes ownership, usage patterns, aliases, and significance. It also includes dynamic updates and corrections, allowing us to test whether agents revise object memory as dialogue changes previously provided information.

\paragraph{Trajectory and observations.}
For each episode, we sample 5--16 navigation goals. Rather than following shortest paths between them, the trajectory visits the goals with intermediate exploration segments.
This produces both explicitly visited goal objects and incidental objects observed along the way.
We record RGB-D frames and agent poses using Habitat-sim~\cite{savva2019habitat}. An object counts as observed when its HM3D semantic mask covers at least 5\% of a rendered view.
For each observed object, we compute episodic memory content, including observation order, observation frequency, and association with specific navigation goals.

\paragraph{Dialogue generation.}
Dialogue is generated in two stages, so that object-level information is fixed before any utterance is written.
We first construct a sequence of structured turn specifications.
Each fixes the speaker, location, turn type, visible and referenced objects, and the memory content the turn expresses, with at most one object-level item per turn to keep grounding verifiable.
An information selection policy restricts which content types a turn may use: perceptual content appears only in navigation-related turns, while social content appears in commentary, recall, and post-arrival turns.
Episodic facts are never stated at all; observation order and frequency must be recovered from the trajectory.
Navigation turns do, however, define the segments that episodic references rely on, as in \textit{``the bed you saw before I asked you to go to the table.''}
Some turns also reference objects observed earlier but not visible at the current waypoint, mirroring natural recall.
Both cases require relating dialogue to the trajectory and to earlier observations.
The LLM then converts each specification into an utterance, conditioned on speaker persona and dialogue context while constrained to the specified content.

\paragraph{Task instantiation.}
Navigation instructions and QA pairs are generated from structured specifications rather than free-form prompts.
For each object, we enumerate the available memory content and construct task specifications that identify the object using one or more memory content types.
Each specification fixes the target object, the target-identification evidence, and the answer evidence before any text is generated; the LLM only converts this specification into natural language.
While target identification may combine multiple content types, each answer is drawn from a single content type.
For QA, we exclude specifications whose target is identified by a single content type and whose answer is drawn from that same type. Otherwise, identifying the object would already reveal the answer.
Because every turn and task item is specified before generation, the benchmark records which object each turn refers to and which content type each item draws on.
This also determines whether each QA answer is stated in dialogue: 60.5\% are verbalized, while the remaining 39.5\% appear only in the visual or trajectory streams (Appendix~K).
These annotations also enable the controlled link-quality and oracle interventions we later use to diagnose memory failures.

\paragraph{Quality control.}
Both dialogue and task items pass through rule-based validation and an LLM-based grounding audit, since LLM utterance generation can introduce content not present in the specification.
Rule-based checks cover metadata leakage, format constraints, information-policy violations, and visibility alignment.
The LLM-based audit checks whether each utterance and task is consistent with its referenced memory content.
Items that fail are regenerated; items still inconsistent after regeneration are dropped.
For QA, we first exclude social-answer questions whose answer evidence is not grounded in the dialogue.
We then remove under-specified items. Within each episode, we group QA and navigation items by object category and retrieval cue. If a group spans multiple distinct objects of the same category, its cue does not uniquely identify the target, and we exclude the entire group.
A final deterministic filter removes QA items whose question already reveals the expected answer (Appendix~A).
Finally, we validate the resulting benchmark with human annotation. Three annotators independently label 450 sampled items: 200 QA, 200 navigation, and 50 dialogue turns. By majority judgment, 98.4\% are valid (Appendix~B).

\subsection{Dataset Statistics}
Table~\ref{tab:stats} summarizes \textsc{DialogMem} after filtering. The benchmark contains 142 episodes from 27 scenes with 10{,}197 dialogue turns, 14{,}193 QA pairs, and 11{,}890 navigation instructions. Episodes average 894.6 RGB-D frames and 71.8 dialogue turns, requiring agents to organize information over long visual and dialogue streams. Both task types cover perceptual, episodic, and social references, and about half the items combine more than one content type (49.5\% of QA and 49.6\% of navigation). For QA, the answer is drawn from a single content type, which may differ from the one used to identify the target.

\begin{table}[t]
\centering\small
\setlength{\tabcolsep}{4pt}
\begin{tabular}{lrr}
\toprule
Target-identification content & QA & Navigation \\
\midrule
Perceptual & 2{,}773 & 2{,}012 \\
Episodic & 1{,}231 & 1{,}531 \\
Social & 3{,}170 & 2{,}446 \\
Compositional & 7{,}019 & 5{,}901 \\
\midrule
Total & 14{,}193 & 11{,}890 \\
\bottomrule
\end{tabular}
\caption{\textbf{\textsc{DialogMem} task composition.} QA and navigation items grouped by the memory content used to identify the target object; compositional items combine two or more content types. Overall scale: 27 scenes, 142 episodes, 296--2{,}405 frames per episode (avg.\ 894.6) with avg.\ 71.8 dialogue turns.}
\label{tab:stats}
\end{table}

\section{Experiments}
We use \textsc{DialogMem} to study how memory representations should organize long posed RGB-D and dialogue streams for embodied tasks.
Our experiments are structured around three questions: (Q1) how do different memory representations perform across question answering and navigation? (Q2) does it matter whether dialogue enters memory during formation or at evaluation time, and does this vary by memory content type? and (Q3) what must a memory representation provide to close the remaining performance gap?

\subsection{Setup}
\paragraph{Baselines.}
We compare nine memory representations grouped into four categories.
All share the same interface. Each constructs a memory $\mathcal{M}_T$ from Phase~1 streams and applies its own retrieval function $\mathcal{R}$ to select evidence $z$ at evaluation time, which a fixed VLM reasoner $\mathcal{F}$ turns into the output. Performance differences across baselines therefore primarily reflect memory formation and retrieval rather than reasoning.

\textbf{Raw stream storage} preserves Phase~1 streams without explicit organization: \textbf{Dialogue} stores all turns as text, \textbf{Frames} stores all RGB-D frames with poses, and \textbf{Frames+Dialogue} stores both.
The reasoner receives the full dialogue when available and 16 uniformly sampled frames. An ablation on Frames+Dialogue up to 128 frames (Appendix~G) shows that a larger budget does not close the gap to structured visual memory.
For navigation, these baselines select one of the sampled frames and use its observation pose. Dialogue is the exception: it stores no frames or poses.

\textbf{Structured visual memory} organizes Phase~1 observations into object-centric structures.
\textbf{ConceptGraph (CG)}~\cite{gu2024conceptgraphs} builds an object-centric memory of detected instances, where each instance is represented by an image crop, an object label, and an estimated 3D location.
\textbf{3D-Mem}~\cite{yang20253d} groups spatially co-located objects into snapshots, each storing a scene image that captures the co-observed objects together with their individual crops.
Following the setting used for both methods in 3D-Mem, the retriever uses the VLM to predict the object categories most relevant to the question or instruction. 
Only the instances (CG) or snapshots (3D-Mem) whose detected categories match are retained as the evidence $z$.
For QA, the reasoner answers from the retained evidence; for navigation, it selects one object instance or snapshot, whose estimated 3D location is used as the predicted destination.

\textbf{Structured visual memory with dialogue} adds the full Phase~1 dialogue as textual evidence at evaluation time, leaving the visual memory and navigation interfaces unchanged (\textbf{CG+Dialogue}, \textbf{3D-Mem+Dialogue}).

\textbf{Structured visual memory with linked dialogue} instead grounds each turn to specific snapshots during formation. At evaluation time, each retrieved snapshot carries only its linked turns, and unlinked turns are dropped.
We instantiate two linkers on 3D-Mem. The \textbf{CLIP linker} links each turn to the snapshots whose scene images are most similar to the turn text under CLIP~\cite{radford2021learning}. The \textbf{VLM linker} uses the VLM to identify which co-observed object the turn refers to, then links the turn to that object's snapshots.
These are deliberately simple baselines rather than proposed methods, probing whether grounding dialogue during formation beats deferring it to evaluation time.

\paragraph{Metrics.}
For QA, we report \textit{LLM-match}~\cite{majumdar2024openeqa}: an LLM judge rates the semantic agreement between predicted and reference answers on a 1--5 rubric, mapped linearly to 0--100 and averaged over items.
Instead of a proprietary model~\cite{hurst2024gpt}, we use an open-source model (Qwen3-VL-4B-Instruct~\cite{bai2025qwen3}, temperature 0) for reproducibility; its scores correlate closely with GPT-4o's (Pearson $r{=}0.87$; Appendix~N).
For navigation, our main metric is \textit{geodesic success}: the predicted target location must lie within 1.0 m geodesic distance of the object's ground-truth location on the scene navigation mesh.
We score this location directly, without path execution, so results reflect spatial grounding in memory rather than planning or actuation.
We additionally report \textit{execution success} (arriving within 1.0 m after rolling out the predicted trajectory) and \textit{SPL} (success weighted by path efficiency; Appendix~Q).

\paragraph{Implementation details.}
We use Qwen3-VL-4B-Instruct~\cite{bai2025qwen3} for all VLM roles and run inference on a single NVIDIA RTX A6000 GPU with 48 GB of memory.
For structured methods, memory construction and 3D location estimation follow the publicly released implementations~\cite{gu2024conceptgraphs, yang20253d}.

\subsection{Results and Analysis}

\begin{table}[h]
\centering
\small
\setlength{\tabcolsep}{4pt}
\begin{tabular}{lcccc}
\toprule
Memory & QA (LLM-m.) & Geo. & Exec. & SPL \\
\midrule
\multicolumn{5}{l}{\textit{Raw stream storage}} \\
Dialogue         & 56.5 & -- & -- & -- \\
Frames           & 24.3 & 11.3 & 10.3 & 10.1 \\
Frames+Dialogue  & 60.1 & 12.7 & 11.5 & 11.3 \\
\midrule
\multicolumn{5}{l}{\textit{Structured visual memory}} \\
CG               & 19.8 & 25.2 & 22.1 & 21.9 \\
3D-Mem           & 35.1 & 31.0 & 30.9 & 30.6 \\
\midrule
\multicolumn{5}{l}{\textit{Structured visual memory with dialogue}} \\
CG+Dialogue      & 58.4 & 25.5 & 22.8 & 22.7 \\
3D-Mem+Dialogue  & 63.4 & 31.1 & 30.1 & 29.7 \\
\midrule
\multicolumn{5}{l}{\textit{Structured visual memory with linked dialogue}} \\
CLIP linker      & 58.8 & 30.6 & 30.3 & 29.9 \\
VLM linker       & 48.6 & 30.8 & 30.4 & 30.1 \\
\bottomrule
\end{tabular}
\caption{\textbf{Results across memory representations.} Geo.\ is geodesic success. Exec.\ is execution success after rolling out the predicted trajectory, and SPL is success weighted by path efficiency.}
\label{tab:main}
\end{table}

\paragraph{Dialogue and visual memory support different tasks (Q1).}
Across the nine memory representations, no single one supports both tasks (Table~\ref{tab:main}).
Dialogue supports moderate QA (56.5) but stores no poses and cannot localize objects at all. 
Frames localizes to some degree (11.3) but is weak on QA (24.3).
Structured visual memory outperforms raw frame storage on navigation but stays below Dialogue on QA.
The two tasks thus draw on different representations, and we next ask whether adding dialogue to visual memory helps both.

\paragraph{Appending dialogue helps QA but leaves navigation ungrounded (Q2).}
Appending the full Phase~1 dialogue at evaluation time yields large QA gains.
3D-Mem improves from 35.1 to 63.4 and Frames from 24.3 to 60.1.
These gains concentrate on answers already stated in the dialogue: 3D-Mem+Dialogue scores far higher on verbalized items than on others, while visual-only 3D-Mem is nearly invariant to this split (Appendix~K).
Episodic answers gain the least from dialogue and are the lowest after appending.
The gains also shrink when a referenced fact has been updated.
On social facts that a later turn updates or corrects, 3D-Mem+Dialogue's QA score drops substantially (77.4 to 61.1, Appendix~M).
Appended dialogue does not separate an object's latest state from its earlier mentions.
Navigation, in contrast, barely moves: 3D-Mem goes from 31.0 to 31.1 (McNemar's test, $p{=}0.42$; Appendix~F) and Frames from 11.3 to 12.7.
One natural explanation is that the reasoner cannot extract useful evidence from appended dialogue.
The same appended dialogue, however, lifts social-answer QA from 20.9 to 76.9 (Table~\ref{tab:qa_content}), so the reasoner can read and use it.
Yet social navigation instructions, whose targets are specified entirely through dialogue, stay at 32.1, as do all other content types (Table~\ref{tab:nav_content}).
The referenced objects are also in memory as detections (Appendix~E).
What appears to be missing is the link between dialogue and the object entries.
Dialogue tells the agent which object is meant and what to answer, but appending it at evaluation time does not attach that evidence to the object entries navigation must localize.

\begin{table}[t]
\centering
\small
\setlength{\tabcolsep}{4pt}
\begin{tabular}{lcccc}
\toprule
Memory & Perc. & Epis. & Soc. & Comp. \\
\midrule
\multicolumn{5}{l}{\textit{Raw stream storage}} \\
Dialogue              &  -- &  -- &  -- &  -- \\
Frames                & 11.3 &  8.8 & 10.9 & 12.1 \\
Frames+Dialogue       & 12.4 & 10.1 & 13.7 & 13.0 \\
\midrule
\multicolumn{5}{l}{\textit{Structured visual memory}} \\
CG                    & 24.9 & 25.2 & 27.0 & 24.6 \\
3D-Mem                & 32.1 & 29.0 & 32.1 & 30.6 \\
\midrule
\multicolumn{5}{l}{\textit{Structured visual memory with dialogue}} \\
CG+Dialogue           & 26.8 & 24.2 & 27.0 & 24.7 \\
3D-Mem+Dialogue       & 32.0 & 29.4 & 32.1 & 30.8 \\
\midrule
\multicolumn{5}{l}{\textit{Structured visual memory with linked dialogue}} \\
CLIP linker           & 32.1 & 28.5 & 31.7 & 30.2 \\
VLM linker            & 32.1 & 28.7 & 31.4 & 30.8 \\
\bottomrule
\end{tabular}
\caption{\textbf{Navigation geodesic success by instruction content type.} Perc., Epis., Soc., and Comp.\ denote perceptual, episodic, social, and compositional instructions.}
\label{tab:nav_content}
\end{table}

\begin{table}[h!]
\centering
\small
\setlength{\tabcolsep}{4pt}
\begin{tabular}{lrrr}
\toprule
Memory & Perc. & Epis. & Soc. \\
\midrule
\multicolumn{4}{l}{\textit{Raw stream storage}} \\
Dialogue            & 72.8 & 27.2 & 72.5 \\
Frames              & 41.3 & 33.5 &  6.5 \\
Frames+Dialogue     & 77.8 & 33.6 & 72.6 \\
\midrule
\multicolumn{4}{l}{\textit{Structured visual memory}} \\
CG                  & 35.1 & 15.0 & 15.2 \\
3D-Mem              & 61.6 & 34.0 & 20.9 \\
\midrule
\multicolumn{4}{l}{\textit{Structured visual memory with dialogue}} \\
CG+Dialogue         & 71.7 & 31.1 & 74.4 \\
3D-Mem+Dialogue     & 77.4 & 38.5 & 76.9 \\
\midrule
\multicolumn{4}{l}{\textit{Structured visual memory with linked dialogue}} \\
CLIP linker         & 71.3 & 36.4 & 70.9 \\
VLM linker          & 67.2 & 35.6 & 49.2 \\
\bottomrule
\end{tabular}
\caption{\textbf{QA LLM-match by answer source.} Columns group QA items by the memory content type the answer is drawn from, marginalizing over the target-identification cue. }
\label{tab:qa_content}
\end{table}

\paragraph{Off-the-shelf linkers do not supply the missing link.}
The natural fix is to create the missing links during memory formation rather than at evaluation time.
Unlike appending, the CLIP and VLM linkers attach each turn to specific snapshots as the memory is built.
Neither improve navigation over 3D-Mem's 31.0 (30.6 and 30.8), and both underperform appending on QA (58.8 and 48.6 vs. 63.4), with the VLM linker well below it (Table~\ref{tab:main}).
This leaves open whether the approach is wrong or the linkers are simply too weak.

\begin{figure}[t]
\centering
\includegraphics[width=0.8\linewidth]{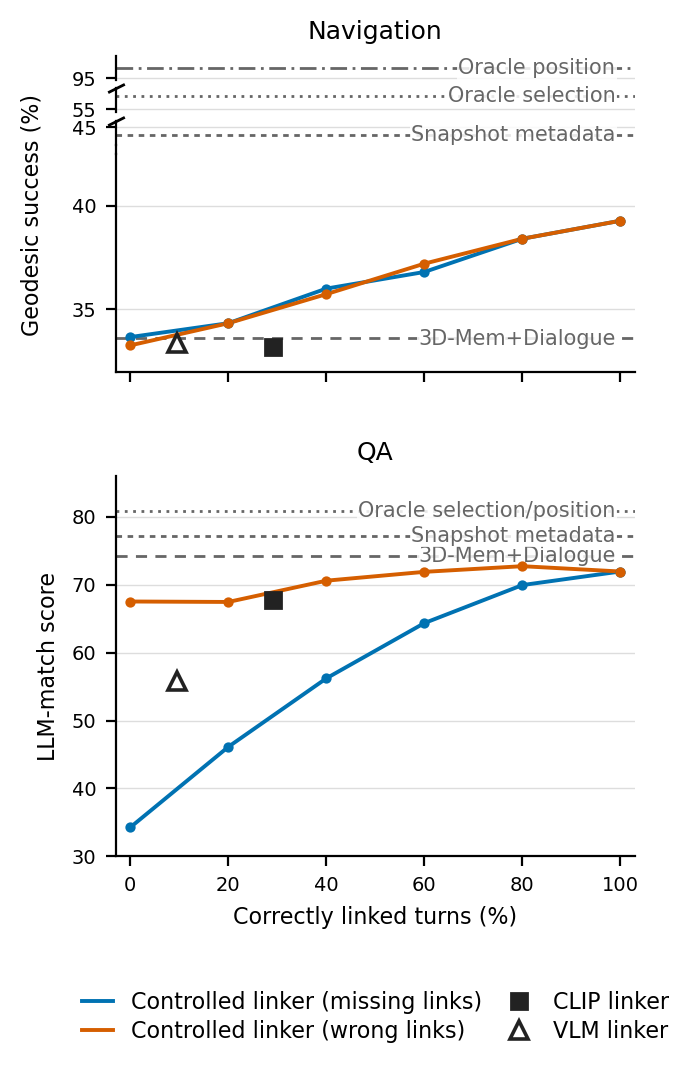}
\caption{\textbf{Downstream effect of dialogue--object link quality.} Curves show a controlled linker swept over link quality under two error modes; markers show the CLIP and VLM linkers, and dashed lines the no-link baseline and the three ground-truth probes. Values are means over three evaluation seeds, with standard deviations at most 0.7 points.}
\label{fig:link_precision}
\end{figure}

\paragraph{Navigation improves with dialogue--object link quality.}
To separate these two possibilities, we control \emph{link quality} directly.
We define it as the percentage of dialogue turns linked to a snapshot that contains the object they refer to, and sweep it from 0 to 100 (Figure~\ref{fig:link_precision}).
The sweep, the ground-truth probes, and the CLIP and VLM linkers share a subset of 500 navigation and 500 QA items, stratified across memory content types. 
These values therefore differ from the full-benchmark numbers in Table~\ref{tab:main}.
At each level, the turns without a correct link are either left unlinked (\emph{missing links}) or attached to a wrong snapshot (\emph{wrong links}), separating the cost of low recall from that of low precision.
Navigation improves steadily with link quality but tops out at 39.3 under perfect links, so linking is the right target yet not sufficient on its own.
The CLIP and VLM linkers sit near the bottom of this curve, correctly linking only 29.1\% and 9.5\% of turns, so better linkers would move navigation up the curve.
On QA the two error modes diverge. Missing links collapse QA (34.3 at link quality 0), while wrong links leave it essentially flat (67.5). QA thus depends on whether the relevant dialogue is present at all, not on which snapshot it is attached to.
This also explains the VLM linker's low QA: it links only 9.5\% of turns, so most of the dialogue never reaches the reasoner.

\paragraph{Closing the remaining gap requires grounding, retrieval, and localization (Q3).}
The dashed reference lines in Figure~\ref{fig:link_precision} decompose the remaining navigation gap with ground-truth probes that keep the 3D-Mem pipeline unchanged.
This gap is not a reasoner limitation.
Our default reasoner (Qwen3-VL-4B) reaches 33.6, the 3D-Mem+Dialogue baseline in Figure~\ref{fig:link_precision}; the larger Qwen3-VL-8B scores 31.5 and GPT-4o 28.8.
The larger reasoners are simply more conservative: they more often answer \emph{unknown} rather than committing to a memory entry, which the benchmark scores as failure (Appendix~H).
For \emph{grounding and organization}, the \textbf{Snapshot metadata} probe attaches each object's ground-truth metadata to the snapshots containing it, including episodic facts that never appear in dialogue. 
This raises navigation from 33.6 to 44.1, above even the perfect-link point (39.3). 
The margin reflects two gains: organizing evidence into object-level facts, and adding episodic content.
For \emph{retrieval}, the model must still select the correct entry from this grounded memory. \textbf{Oracle selection} replaces that choice with the ground-truth object's memory entry while keeping 3D-Mem's estimated position, lifting navigation to 56.6.
For \emph{localization}, \textbf{Oracle position} further replaces the estimated pose with the object's ground-truth location, raising navigation to 96.2. 
The selection-to-position gap is thus dominated by error in 3D-Mem's estimated 3D locations.
Both oracle probes still require the reasoner to confirm that the supplied entry matches the instruction, answering yes or unknown, so they isolate a single memory component rather than granting success automatically.
The small shortfall of oracle position from a perfect score is the reasoner occasionally rejecting even the correct entry, not a memory error.
Together, the decomposition points to three targets: grounding dialogue to the right objects and organizing it into object-level facts during memory formation, retrieving the correct entry at evaluation time, and estimating its 3D location accurately.

\section{Conclusion}
This paper introduces \textsc{DialogMem}, a benchmark for studying how embodied memory should organize what an agent observes together with what humans tell it.
\textsc{DialogMem} provides time-aligned posed RGB-D and dialogue streams for memory formation, and navigation and QA tasks that query the resulting memory over perceptual, episodic, and social content.
Evaluating nine memory representations with controlled diagnostics, we isolate three research directions..
First, memory formation should ground what the agent observes and what it is told around the same object entries.
Second, retrieving the correct entry remains a substantial bottleneck, even when ground-truth facts are attached to the correct entries.
Third, object locations must be estimated accurately; this is the limiting factor once the correct entry is retrieved.
\textsc{DialogMem} provides a testbed for measuring progress toward embodied agents that remember and act on what humans tell them over time.

\noindent\textbf{Limitations.}
\textsc{DialogMem} uses LLM-generated dialogue derived from structured object-level specifications.
While this enables exact grounding annotations and controlled diagnostics, it does not capture the full ambiguity, noise, and dynamics of spontaneous human dialogue.
The benchmark uses 27 real-world HM3D scans but evaluates agents in simulation with noise-free poses, so the findings may not fully transfer to broader environment types or physical robots with perception and pose errors.
Our study diagnoses existing memory representations rather than proposing new ones. The oracle probes establish upper bounds on each component's contribution, but designing a memory representation that jointly realizes these gains is left to future work. Our analysis is instantiated with 3D-Mem and a small set of VLM reasoners; whether our findings generalize to other memory architectures, linkers, and reasoners remains to be validated.


\section*{Acknowledgments}
This work was partly supported by grants from the IITP
(RS-2021-II211343-GSAI/10\%, RS-2022-II220951-LBA/10\%,
RS-2022-II220953-PICA/10\%, RS-2026-25553157-MIACC/10\%),
the NRF (RS-2024-00353991-SPARC/15\%, RS-2023-00274280-HEI/15\%),
the KEIT (RS-2025-25453780/15\%), and the KIAT
(RS-2025-25460896/15\%), funded by the Korean government.

\bibliography{aaai2027}


\end{document}